\documentclass{ifacconf}
\usepackage{graphicx}      
\usepackage{natbib}        
\usepackage{amsmath}
\usepackage{amsfonts}
\usepackage{amssymb}

\usepackage{color}
\usepackage{algorithm}
\usepackage{algpseudocode}

\RequirePackage[absolute]{textpos}
\DeclareRobustCommand*{\copyrightnote}[1]{%
  \begin{textblock}{200}(15,286.2)
      \footnotesize #1%
  \end{textblock}%
}

\newtheorem{proof}{Proof}

\AtBeginEnvironment{thebibliography}{\linespread{0.95}\selectfont}

\newcommand{\Vt}{\frac{\partial V}{\partial t}}
\newcommand{\Vx}{\frac{\partial V}{\partial x}}

\begin{document}
\begin{frontmatter}

\title{%
Learning optimal controllers: a dynamical motion primitive approach\thanksref{footnoteinfo}%
} 

\thanks[footnoteinfo]{%
This project has received funding from the European Union's Horizon 2020 research and innovation programme under the Marie Skłodowska-Curie grant agreement no. 899987. This work was also supported by the German Research Foundation (DFG) as part of Germany’s Excellence Strategy, EXC 2050/1, Project ID 390696704 – Cluster of Excellence “Centre for Tactile Internet with Human-in-the-Loop” (CeTI) of Technische Universität Dresden.  
The authors would also like to thank Luis F.C. Figueredo and Dennis Ossadnik for the scientific discussions. %
}

\author[First]{Hugo T. M. Kussaba}
\author[First,Second,Fourth]{Abdalla Swikir}
\author[First]{Fan Wu}
\author[First]{Anastasija Demerdjieva}
\author[Third]{Gitta Kutyniok}
\author[First,Fourth]{Sami Haddadin}

\address[First]{Chair of Robotics and Systems Intelligence, Munich Institute of Robotics and Machine Intelligence, Technical University of Munich, Germany.}
\address[Second]{Department of Electrical and Electronic Engineering, Omar Al-Mukhtar University, Libya.}
\address[Third]{Mathematical Institute, Ludwig-Maximilians University of Munich, Germany.}
\address[Fourth]{Centre for Tactile Internet with Human-in-the-Loop (CeTI), Germany.}

\begin{abstract}                %
Real-time computation of optimal control is a challenging problem and, to solve this difficulty, many frameworks proposed to use learning techniques to learn (possibly sub-optimal) controllers and enable their usage in an online fashion.
Among these techniques, the optimal motion framework is a simple, yet powerful technique, that obtained success in many complex real-world applications. 
The main idea of this approach is to take advantage of dynamic motion primitives, a widely used tool in robotics to learn trajectories from demonstrations. While usually these demonstrations come from humans, the optimal motion framework is based on demonstrations coming from optimal solutions, such as the ones obtained by numeric solvers.

As usual in many learning techniques, a drawback of this approach is that it is hard to estimate the suboptimality of learned solutions, since finding easily computable and non-trivial upper bounds to the error between an optimal solution and a learned solution is, in general, unfeasible. 
However, we show in this paper that it is possible to estimate this error for a broad class of problems. Furthermore, we apply this estimation technique to achieve a novel and more efficient sampling scheme to be used within the optimal motion framework, enabling the usage of this framework in some scenarios where the computational resources are limited.
\end{abstract}


\end{frontmatter}

\copyrightnote{\copyright 2023 the authors. This work has been accepted to IFAC for publication under a Creative Commons Licence CC-BY-NC-ND.}

\section{Introduction}

Optimal control problems (OCPs) arise in many applications in robotics, ranging from motion planning \citep{LaValle2006Planningalgorithms} to robot design \citep{DinevEtAl2022DifferentiableOptimalControl}.
Generally, analytical solutions are not known for these problems, which then must be solved by numerical methods such as direct methods \citep{DiehlEtAlFastDirectMultiple2006}.
These methods transcribe the optimal control problem into a non-linear programming (NLP) problem that can be usually solved by interior-point methods even for non-convex problems  \citep{NocedalEtAlAdaptiveBarrierUpdate2009}. 
Nevertheless, many problems in robotics result in an NLP with a large number of variables and constraints. Hence, finding numerical solutions for those problems requires extensive computational power which in turn makes the usage of OCP solvers intractable for high dimensional systems (such as robotics systems with many degrees of freedom), especially in real-time. 

Common ways to overcome this restriction and enable the online usage of optimal controllers are based on exploiting \textit{a priori} information of the solution \citep{HarzerEtAlEfficientNumericalOptimal2022} and/or by providing near-optimal solutions.
The latter approach, for instance, has been explored several times in model-predictive controllers (MPC), where embedded applications with low computational power or limited amount of energy prohibit solving the nonlinear optimization problem online
\citep{PohlodekEtAlFlexibledevelopmentevaluation2022}.

In view of this, machine-learning based techniques which exploit the structure of the problem to yield approximate optimal controllers are a natural candidate for obtaining near-optimal solutions in real-time.
Indeed, the intersection between machine learning and optimization problems has shown promising results ranging from 
combinatorial optimization problems \citep{BengioEtAlMachinelearningcombinatorial2021},
MPC controllers \citep{WeinanEtAlEmpoweringOptimalControl2022, KargLuciaEfficientRepresentationApproximation2020}
and solving Hamilton-Jacobi-Bellman partial differential equation \citep{BansalTomlin2021DeepReachDeepLearning}.

In particular, some supervised learning methods leverage the numerical solution of optimal control problems to generate training data for teaching a function approximator to the optimal control solution. This ``Learning from Optimal Control'' (LfOC) approach has been successfully demonstrated for real-time quadrotor aggressive maneuvers in \cite{TomicEtAl2014Learningquadrotormaneuvers} and throwing/reaching tasks for robotic manipulators in \cite{HaddadinEtAl2013OptimalControlViscoelastic}, where dynamical motion primitives (DMPs), also referred to as dynamical movement primitives, act as the function approximator for the optimal control solution.

While the aforementioned LfOC methods are not burdensome when used online, the generation of training datasets can be very time-expensive since it requires solving multiple OCPs. %
Moreover, missing an important sample in the training dataset could lead to very costly solutions.
Furthermore, there are no results so far on defining a metric to measure the performance of learned controllers compared to the optimal ones.
Thus, it is a practical concern how to define a metric that quantifies the error between the out-of-sample solutions and the real optimal solution, and establishes a framework to choose the most representative solutions for decreasing such an error.

In this work, we develop the basis for such a framework by leveraging an important information obtained from the training trajectories: their optimal costs and the sensitivity of the optimal value function corresponding to the initial condition of these trajectories. Based on this information, we devise a way to estimate the suboptimality of out-of-sample solutions and, moreover, show how to take advantage of this estimate for efficiently selecting training data.  

\section{Preliminaries}

\subsection{Optimal control problems}
In this paper, we consider systems of the form
\begin{equation}\label{eq:affine_control_sys}
\dot{x}(t) = f(x(t))+g(x(t)) u(t)
\end{equation}
where $x \in \mathbb{R}^n$ is the state, $u \in \mathbb{R}^m$ is the control input, and $f: \mathbb{R}^n \rightarrow \mathbb{R}^n$ and $g: \mathbb{R}^n \rightarrow \mathbb{R}^{n \times m}$ 
are smooth functions describing the system dynamics. 

Motion planning problems linked to \eqref{eq:affine_control_sys} are frequently 
cast as an optimal control problem of the form
\begin{subequations}\label{eq:OCP}
    \begin{equation}\label{eq:OCP_cost}
        \underset{x \in \mathcal{X}, u \in \mathcal{U}}{\operatorname{minimize}} \int_0^{t_f} L(x(t), u(t)) \, d t
    \end{equation}
subject to
    \begin{align} \label{eq:constrained_sys}
        \dot{x}(t) &= f(x(t)) + g(x(t))u(x(t))), \,\, t \in [0,t_f], \\ 
        x(0) &= x_0,\,\, x(t_f) = x_f, \nonumber
    \end{align}
\end{subequations}
where $\mathcal{X}$ is the set of differentiable functions $x:[0, t_f]
\rightarrow \mathbb{R}^n$, %
$\mathcal{U}$ is the set of continuous functions $u:[0, t_f]
\rightarrow \mathbb{R}^m$ with $u(0) = 0$, and $L:\mathcal{X} \times \mathcal{U} \rightarrow \mathbb{R}$ is the running cost, given by
$$
L(x(\cdot), u(\cdot)) = Q(x) + u^\mathsf{T} R u,
$$
with $Q:\mathbb{R}^n \rightarrow \mathbb{R}$ either identically to zero, or a continuous and positive definite function,
and $R \in \mathbb{R}^{m \times m}$ is a positive-definite symmetric matrix.
Moreover, $x_0 \in \mathbb{R}^n$ is a given initial state, $x_f \in \mathbb{R}^n$ and $t_f > 0$ are respectively the terminal state and terminal time. 

We assume that \eqref{eq:OCP} has a unique solution for each $x_f \in \mathbb{P}$, where $\mathbb{P} \subseteq \mathbb{R}^n$ is a compact set, and denote the optimal state trajectory and optimal control respectively by $x^*$ and $u^*$.
Many problems in motion planning require solving (or at least finding an approximate solution) of \eqref{eq:OCP} for multiple instances. In particular, given a compact set $\mathbb{P}$ with the desired points to be reached, we would need to solve one instance of \eqref{eq:OCP} for each fixed $x_f \in \mathbb{P}$.

In many real-world applications, it is often impossible to find a closed-form solution for \eqref{eq:OCP}, and numerical methods that transform the optimal control problem into a non-linear programming problem are widely used \citep{DiehlEtAlFastDirectMultiple2006}.
Those methods, known as direct methods, are based on the time discretization of the state and control variables. Thus, the obtained solution from the NLP is generally inferior to the solution of the original OCP and a finer discretization may be needed to improve the solution. The price of making the discretization finer is increasing the cost of optimization by requiring more iterations to be performed within the larger search space \citep{SahlodinBartonEfficientControlDiscretization2017}. 
Consequently, one can not use direct methods in real-time in general. Indeed, even if $\mathbb{P}$ is a finite set, it can be very time-consuming to produce optimal solutions that cover the entire set $\mathbb{P}$. 

\subsection{Review of DMP-based optimal motion framework}

Since in general it is not possible to obtain a closed form solution to \eqref{eq:OCP} and the complexity of numerical methods limits their usage in real-time, many works in the literature propose to leverage learning techniques to obtain near-optimal solutions in real time.
In this context, \cite{HaddadinEtAl2013OptimalControlViscoelastic} proposed a learning framework for optimal control based on dynamical motion primitives (DMPs). %
Before reviewing this framework, we will briefly review the theory of DMPs and the features that make them an attractive tool within the context of a learning optimal control framework.  
For more in-depth details of DMPs, the reader is referred to \cite{SaverianoEtAlDynamicMovementPrimitives2021}.

\subsubsection{Dynamic movement primitives}

Following the notation of \cite{WeitschatEtAl2013Dynamicoptimalityreal}, a DMP is a dynamical system defined by the differential equation
\begin{equation}\label{eq:DMP_definition}
    -\tau^{2}\ddot{x}(t)+\kappa(x_f -x(t))-D\tau\dot{x}(t)= F(s(t))
\end{equation}
where $\tau > 0$, $\kappa > 0$ and $D > 0$ are tuning parameters, and $F:\mathbb{R} \rightarrow \mathbb{R}^n$ and $s: \mathbb{R} \rightarrow \mathbb{R}$ are functions defined as
\begin{align}
    F_i(s(t)) &= \frac{\sum_{j=1}^{N}\omega_{i,j}\,e^{-h_{i,j}(s(t)-c_{i,j})^{2}}}{\sum_{j=1}^{N}e^{-h_{i,j}(s(t)-c_{i,j})^{2}}}s(t),
    \label{eq:DMP_forcing_function} \\ 
    s(t) &= e^{-(\alpha/\tau)t}, \label{eq:DMP_clock}
\end{align}
with $\alpha > 0$. The forcing function $F$ is a sum of $N$ Gaussian basis functions with center points $c_{i,j}$ and widths $h_{i,j}$ given by
$$
c_{i,j}= e^{\left(-\alpha \frac{j-1}{N-1}\right)},
$$
$$
h_{i,j}= \begin{cases}{(c_{i,j+1}-c_{i,j})^{-2}}, & j=1, \ldots, N-1, \\ h_{i,N-1} & j=N, \end{cases}
$$
and the weights $\omega_{i,j}$ are learned from data by minimizing the error given by
\begin{equation}\label{eq:error_DMP_learning}
\sum_{k} \| -\tau^{2}\ddot{x}(t_k)+\kappa(x_f -x(t_k))-D\tau\dot{x}(t_k) - F(s(t_k)) \|,    
\end{equation}
where $x(t_k)$, $\dot{x}(t_k)$ and $\ddot{x}(t_k)$ are obtained by sampling a given trajectory with starting point at 
$x(0)=x_0$ and ending point in $x(\tau) = x_f$. 
By the next theorem, solutions of \eqref{eq:DMP_definition} will always converge to the latter point.

\begin{thm}\label{thm:stability_of_goal}
\citep[Sec. 2.1.3]{Ijspeert_2013} 
Suppose that $\kappa = D^2/4$. Then the parameter $x_f$ in  \eqref{eq:DMP_definition} is a global asymptotic stable equilibrium point.
\end{thm}

After learning the weights $\omega_{i,j}$ from a single trajectory $\tilde{x}:\mathbb{R} \rightarrow \mathbb{R}^n$ it is possible to use \eqref{eq:DMP_definition} to generate new trajectories that are qualitatively similar %
to $\tilde{x}$, but with different ending points. In exact terms, to reproduce a new trajectory with ending point in $x_f'$, one integrates the differential equation
\begin{equation}\label{eq:DMP_generalization}
    -\tau^{2}\ddot{x}(t)+\kappa(x_f' -x(t))-D\tau\dot{x}(t)= F(s(t)),\,\,x(0)=x_0.
\end{equation}
It is clear that the only difference between \eqref{eq:DMP_definition} and \eqref{eq:DMP_generalization} are the parameters $x_f$ and $x_f'$. Also note that the same forcing term $F$ learned in \eqref{eq:DMP_definition} is used without changes and there is no need to re-learn it. The convergence of the new trajectory to $x_f'$ is guaranteed by Theorem~\ref{thm:stability_of_goal}.

For distinguishing between these trajectories more easily, we will introduce the following notation: the trajectory obtained by solving \eqref{eq:DMP_definition} with initial condition $x(0) = x_0$ will be denoted as $\Phi_\textrm{D}(x_f\mid x_f)$, while the trajectory obtained by solving \eqref{eq:DMP_generalization} as 
$\Phi_\textrm{D}(x_f'\mid x_f)$.

Upper bounds to the distance between $\Phi_\textrm{D}(x_f\mid x_f)(t)$ and $\Phi_\textrm{D}(x_f'\mid x_f)(t)$ at each time $t$ can be quantitatively expressed as a function of the distance between $x_f$ and $x_f'$. This follows from the fact that the differential equation in \eqref{eq:DMP_definition} has a continuous dependence on the parameter $x_f$ \citep{Perko2001DifferentialEquationsDynamical}. These bounds, however, depend on the computation of (local) Lipschitz 
constants, whereas 
in this work we show that is possible to obtain an easy computable expression for the distance between $\Phi_\textrm{D}(x_f\mid x_f)$($t$) and $\Phi_\textrm{D}(x_f'\mid x_f)(t)$ in terms of the DMP parameters and the distance between $x_f$ and $x_f'$:
\begin{prop}\label{thm:upper_bound_DMP_traj}
The following equality is valid for all $t~\in~[0, t_f]$:
\begin{gather*}
    \|\Phi_\textrm{D}(x_f\mid x_f)(t)-\Phi_\textrm{D}(x_f'\mid x_f)(t)\| = \\
       \left|e^{-\frac{D t}{2 \tau }} \left(-\frac{D t}{2 \tau }-1\right)+1\right| \|x_f-x_f'\|.
\end{gather*}
\end{prop}
\begin{proof}
See Appendix.
\end{proof}

\subsubsection{Learning controllers with optimal motion primitives}

The DMP-based framework relies on the assumption that optimal trajectories that have close terminal end-points have qualitatively similar shapes. While this assumption seems very restrictive, it has been verified in practice for reaching/tracking motions with minimal energy in elastic robots in \cite{WeitschatEtAl2013Dynamicoptimalityreal} and time-optimal maneuvers for quadrotors in \cite{TomicEtAl2014Learningquadrotormaneuvers}.

Under this assumption, the optimal trajectories obtained by numerically solving \eqref{eq:OCP} will be used to create DMPs with different $x_f$'s at each point of a finite grid $\hat{\mathbb{P}}$ contained in $\mathbb{P}$. If a terminal state is outside $\hat{\mathbb{P}}$, it should be inside a hypercube with vertices in $\hat{\mathbb{P}}$, and a trajectory can be computed almost instantaneously by ``blending'' the weights of the DMPs in the vertices. For instance, \cite{TomicEtAl2014Learningquadrotormaneuvers} blends the weights by bilinear interpolation while \cite{HaddadinEtAl2013OptimalControlViscoelastic} uses a weighted interpolation based on the costs of the trajectories. The new DMP is then used to instantaneously generate a new trajectory to the unsampled terminal state.

Assuming that there exists a well-defined function 
\begin{equation}\label{eq:recover_input}
    u := u(x,\dot{x}, \ddot{x}, \ldots, x^{(m)}),
\end{equation}
where $x^{(m)}$ is the $m$-th derivative of $x$ with respect to time, the control input $u$ associated with a trajectory $x$ can be directly recovered from the DMP trajectories.
In particular, if $g$ in \eqref{eq:affine_control_sys} is invertible, %
then $u$ can be recovered with
$$
u(t) = g^{-1}(x(t)) (\dot{x}(t) - f(x(t)))
$$
While in general the new solutions will not be optimal, they will be close to it and the stability of the terminal state is guaranteed by Theorem~\ref{thm:stability_of_goal}. 

Finally, it is interesting to highlight that many extensions of the DMP framework can be inherited by this optimal motion framework. For instance, \cite{TomicEtAl2014Learningquadrotormaneuvers} used the possibility of combining multiple DMPs to join a sequence of maneuvers for a quadrotor.

\section{Estimation of Optimal Value and Sampling Algorithm}

The DMP-based approach for learning optimal control requires storing $KN$ weights in memory, where 
$N$ is the number of basis functions used in \eqref{eq:DMP_definition} and $K$ is the number of sample trajectories obtained by solving \eqref{eq:OCP} for different $x_f \in \mathbb{P}$.
While \cite{WeitschatEtAl2013Dynamicoptimalityreal} proposed a technique for finding a minimal number $N$ of basis functions, the product $KN$ can still be large depending on the requirements of the application.
Moreover, some complex trajectories may require a high number $K$ to obtain good generalization properties, such as time-optimal perching maneuvers in quadrotors \citep{TomicEtAl2014Learningquadrotormaneuvers}. 

To reduce $K$, one may consider to avoid sampling trajectories whenever the cost of the out-of-sample DMP is good enough for a given application. For that, one needs to compute (or at least estimate) how suboptimal a learned trajectory is.
To achieve this cost-aware sampling strategy, we propose to leverage the sensitivity information in the learned trajectories, i.e. the trajectories that are obtained by solving \eqref{eq:OCP}. %

In the next subsection we show how this can be done and how it can be used to estimate the value function in a neighborhood of the ending points of the learned trajectories. We also show how to apply this approximation to efficiently create a sampling grid for learning optimal control when there are restrictions on the number of samples.

\subsection{First-order approximation to the value function}

In order to estimate the suboptimality of an out-of-sample trajectory $\Phi_\textrm{D}(x_f'\mid x_f)$, we will need to estimate the cost of the true optimal trajectory. To this end, we will get the cost of optimal trajectories by using the notion of the value function, whose definition we precisely state next.

Following p.~159 of \cite{liberzon2011calculus}, the value function $V:\mathbb{R}^n \times \mathbb{R} \rightarrow \mathbb{R}$ for \eqref{eq:OCP} is defined as
\begin{equation}\label{eq:optimal_value_def}
V(x,t)=\inf_{u \in \mathcal{U}} \left\{ \int_{t}^{t_{f}}L(x(s),u(s))\,\,ds \right\}.
\end{equation}
In other words, assuming the existence of an optimal control, $V(x,t)$ is the cost of the optimal trajectory that starts from the initial state $x$ at the time $t$. %

Assuming that $V$ is differentiable on $\mathbb{P}$, it is possible to use Bellman's optimality principle to show that $V$ is the solution of a partial differential equation (known as the Hamilton-Jacobi-Bellman equation) and that it is possible to recover the solution of $\eqref{eq:OCP}$ from this $V$. This is precisely stated
for systems of the form \eqref{eq:affine_control_sys} as follows.
Suppose that  \eqref{eq:affine_control_sys} is Lipschitz continuous and stabilizable on $\Omega \subseteq \mathbb{R}^n$, and that $f(0)=0$. Assume that there exists a differentiable function $V:\mathbb{R}^n \times \mathbb{R} \rightarrow \mathbb{R}$ satisfying the partial differential equation given by\footnote{%
Similar to \cite{Isidori} and \cite{Schaft}, we have assumed the existence of differentiable solutions to \eqref{eq:HJB}. We refer the interested readers to \cite[Remark 1]{ChengEtAl2007FixedFinalTime} for a discussion about the differentiability of the function $V(x,t)$.   
}
\begin{equation}\label{eq:HJB}
    \!\!\Vt(x,t) \!=\! - \Vx(x,t)f(x) \! + \!\frac{1}{4}\!\left\| R^{-\frac{1}{2}} \Vx^\mathsf{T}\!(x,t) g(x) \right\|^2\!\!\!.
\end{equation} 

Then the optimal control solving \eqref{eq:OCP} is given by
\begin{equation}\label{eq:input_OCP_sol}
    u^*(x,t) = -\frac{1}{2} R^{-1} g^\mathsf{T}(x) \Vx(x,t).
\end{equation}

Given an optimal trajectory $u^*$ obtained by numerically solving \eqref{eq:OCP} and assuming \eqref{eq:HJB} holds, %
it is possible to recover $\Vx$ along this optimal trajectory:
\begin{equation}\label{eq:estimate_Vx}
    \Vx(x^*(t),t) = -2 g^{-\mathsf{T}}(x^*(t)) R u^*(x^*(t),t).
\end{equation}

In particular, $\Vx(x^*(0), 0)$ can be interpreted as the sensitivity of the optimal cost $V$ with respect to changes in the initial condition $x^*(0)$---for more details, the reader is referred to \citep{KamienSchwartz2012Dynamicoptimizationcalculus}. 
In the DMP-based framework, however, the initial state is fixed while the terminal state $x_f \in \mathbb{P}$ is changed. 
To enable the use of \eqref{eq:estimate_Vx} in our framework, we will use Lemma~\ref{lem:reverse_sys} below.

\begin{lem}\label{lem:reverse_sys}
The optimal cost of the OCP
\begin{subequations}\label{eq:reverse_OCP}
    \begin{equation}\label{eq:reverse_OCP_cost}
    \underset{z(\cdot), v(\cdot)}{\operatorname{minimize}} \int_0^{t_f} L(z(t), v(t)) \, d t %
    \end{equation}
    subject to
    \begin{align} \label{eq:reverse_constrained_sys}
        \dot{z}(t) &= -f(z(t)) - g(z(t))v(t), \,\, t \in [0,t_f], \\ 
         z(0) &= x_f, \,\, z(t_f) = x_0 \nonumber
    \end{align}
\end{subequations}
is equal to the optimal cost of \eqref{eq:OCP}, and the optimal functions are related by
\begin{equation}\label{eq:reverse_time_symmetry}
    z^*(t) = x^*(t_f - t), \,\, v^*(t) = u^*(t_f - t).
\end{equation}
\end{lem}
\begin{proof}
Define the functions $z$ and $v$ as
$$
z(t) = x(t_f - t), \,\, v(t) = u(t_f - t).
$$
By definition, $z(0) = x_f$ and $z(t_f) = x_0$ if and only if $x(0) = x_0$ and $x(t_f) = x_f$.

Moreover, since $\dot{z}(t) = - \dot{x}(t_f - t)$, we have by \eqref{eq:affine_control_sys} that
\begin{align*}
\dot{z}(t) &= - f(x(t_f - t)) - g(x(t_f - t))u(t_f-t) \\
           &= - f(z(t)) - g(z(t))v(t).
\end{align*}
Finally, performing integration by substitution we can show that the cost of the trajectories is the same:
\begin{align*}
\int_{0}^{t_f}  L(x(t), u(t))\,dt &= \int_{t_f}^{0}  -L(x(t_f - t), u(t_f - t))\, dt \\
                                  &= \int_{0}^{t_f}  L(z(t), v(t))\, dt. 
\end{align*}

Thus, solutions of \eqref{eq:constrained_sys} can be mapped to solutions of \eqref{eq:reverse_constrained_sys} with same cost, and by similar arguments, the converse also holds. In particular, the optimal cost of the two OCP problems is the same. \qed
\end{proof}

Lemma~\ref{lem:reverse_sys} allows us to convert a (parametric) optimal control problem in which the initial condition is fixed and the terminal condition is varying in $\mathbb{P}$ to a problem in which the initial condition changes in $\mathbb{P}$ while the terminal condition is fixed. How this helps us to achieve first-order estimates for the optimal value function is explained next. 

First, we solve\footnote{Please note that the we have assumed that the optimal solution is obtained numerically. However, and without losing generality, we only need to assume that an optimal solution is given to us.} the optimal control problem \eqref{eq:reverse_OCP}, obtaining a backward-time trajectory $z^*$ and its associated control input $v^*$. In particular, we obtain the value of $v^*$ which starts at time $0$ in $x_f$, which will be used in the final step.

Second, we define $x^*$  using \eqref{eq:reverse_time_symmetry} and encode the forward-time trajectory solution in the DMP so we can generate a trajectory $\Phi_\textrm{D}(x_f'\mid x_f)$ to a new goal $x_f' \in \mathbb{P}$.

Finally, to estimate the suboptimality of the new out-of-sample trajectory,
we will need to consider the value function associated with the backward optimal control problem \eqref{eq:reverse_OCP}, which is defined as
$$
\tilde{V}(x,t):=\inf_{u \in \mathcal{U}} \left\{ \int_{t}^{t_{f}}L(x(t_f - s), u(t_f - s))\,\,ds \right\},
$$
and note that $\tilde{V}(x_f', 0)$ is, by Lemma~\ref{lem:reverse_sys}, the cost of the optimal trajectory with same endpoints as $\Phi_\textrm{D}(x_f'\mid x_f)$.

To compute $\tilde{V}(x_f', 0)$, note that the differentiability of $V$ (with respect to $x$) implies (see Corollary 1.24 of \cite{guler2010foundations}) 
\begin{equation}\label{eq:first-order-approx}
    \!\!\tilde{V}(x_f', 0) \!=\! \tilde{V}(x_f,0) + \frac{\partial \tilde{V}}{\partial x}(x_f, 0)(x_f' - x_f) \!+\! {o}(\| x_f' - x_f \|),\!\!
\end{equation}
where ${o}(\| x_f' - x_f \|)$ is an (unknown) function that converges to $0$ as $\| x_f' - x_f \|$ tends to $0$.

\begin{figure}
    \centering
    \includegraphics[scale=0.42]{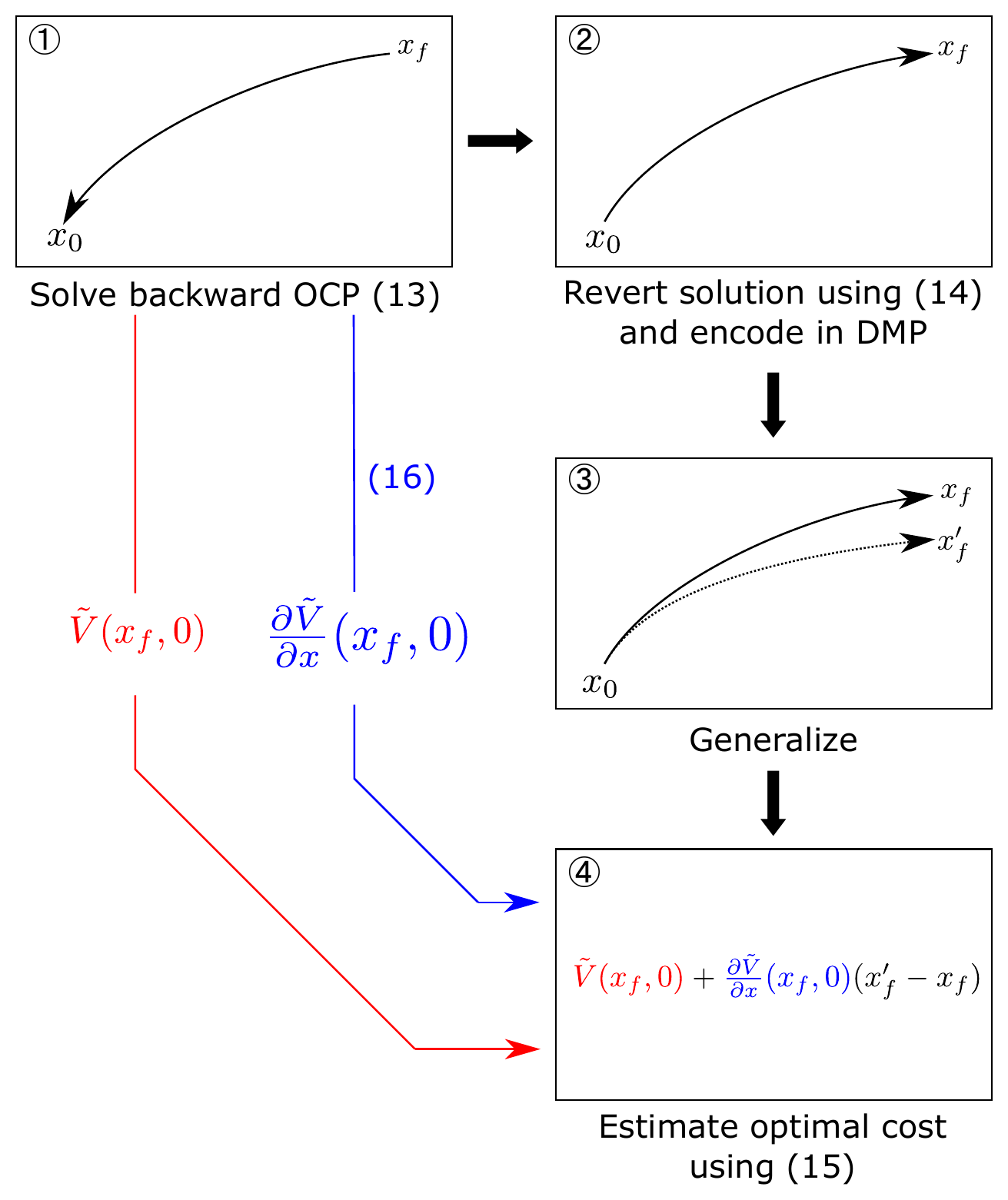}
    \caption{%
    Diagram illustrating the steps for computing an estimate for the cost of an optimal trajectory that corresponds to an out-of sample trajectory with same endpoints.
    }\label{fig:diagram}
\end{figure}

Moreover, the term $\tilde{V}(x_f,0)$ of the right-hand side of \eqref{eq:first-order-approx} 
is already computed by the first step, and the term $\frac{\partial \tilde{V}}{\partial x}(x_f, 0)$ can be computed applying $\eqref{eq:estimate_Vx}$ for the backward-time system, that is\footnote{Note that here we use g instead of -g because we are applying $\eqref{eq:estimate_Vx}$ to the backward-time system.}
\begin{equation}\label{eq:compute_backwards_grad_V}
\frac{\partial \tilde{V}}{\partial x}(x_f,0) = 2 g^{-\mathsf{T}}(x_f) R v^*(x_f,0),
\end{equation}
where $v^*(x_f,0)$ was also computed in the first step. Figure~\ref{fig:diagram} summarizes the above steps.

In conclusion, \eqref{eq:first-order-approx} 
enables us to estimate (for sufficiently small values of $\|x_f' - x_f\|$) the suboptimality of $\Phi_\textrm{D}(x_f'\mid x_f)$ by computing 
\begin{equation}
\begin{gathered}\label{error}
    \int_0^{t_f} L(\Phi_\textrm{D}(x_f'\mid x_f)(s), \hat{u}_{D}(s))\,\,ds \\ - \left[\tilde{V}(x_f,0) + \frac{\partial \tilde{V}}{\partial x}(x_f, 0)  (x_f' - x_f) \right],
\end{gathered}
\end{equation}
where $\hat{u}_{D}$ is the input associated to $\Phi_\textrm{D}(x_f'\mid x_f)$ and is computed by \eqref{eq:recover_input}.

\subsection{Sampling algorithm description}

In this section, we explain how the approximation obtained in \eqref{error} can be used to design a sampling algorithm for applications in which memory storage restrictions possibly prohibit the usage of a fine grid for $\mathbb{P}$.

The details of the sampling scheme are presented in the pseudo-code of Algorithm~\ref{alg:sampling}. First, a terminal point $x_f \in \mathbb{P}$ is chosen and a direction $v \in \mathbb{R}^n$ is given. Then, the backward optimal control with $z(0) = x_f$ is solved and encoded in $\Phi_\textrm{D}(x_f\mid x_f)$.

After that, we select another point in $\mathbb{P}$ by adding a user-defined $\Delta x \in \mathbb{R}^n$ to the starting point. If this new point $x'_f$ is outside $\mathbb{P}$, it means that we do not need to continue any further in this direction and the algorithm stops. 

Otherwise, we use the procedure outlined in the last subsection to estimate the suboptimality of the generalization $\Phi_\textrm{D}(x_f'\mid x_f)$ using \eqref{error}. If the difference between the cost of $\Phi_\textrm{D}(x_f'\mid x_f)$ and the estimated cost of the optimal trajectory ending at $x_f'$ is lower than a user-defined threshold, then it means that the generalization $\Phi_\textrm{D}(x_f'\mid x_f)$ is good enough for the application and there is no need to solve a new optimal control problem for encoding it in a new DMP.

It is important to remark that since \eqref{eq:first-order-approx} is only a first-order approximation, the estimate can become useless if $x_f'$ and $x_f$ are distant. Because of this, the algorithm should re-sample in the case the number of steps given without re-sampling surpass $t_\text{steps}$, a user-defined limit.
The entire procedure repeats until a number $t_\text{samples}$ of samples, defined by the user to reflect the limitations of his or her application, are reached.

By repeating the sampling algorithm for each direction orthogonal to $v$ the algorithm can be used to create a $n$-dimensional grid $\hat{\mathbb{P}}$, whose points are made by the Cartesian product of the samples generated by each direction orthogonal to $v$.

Finally, a near optimal trajectory can be generated online for each point in $\mathbb{P}$ by using a bilinear or a cost-weighted interpolation in $\hat{\mathbb{P}}$  \citep{TomicEtAl2014Learningquadrotormaneuvers, HaddadinEtAl2013OptimalControlViscoelastic}. In either case, it is possible to use \eqref{error} again to estimate the suboptimality of each point $x'_f$ in the entire region $\mathbb{P}$ by using \eqref{error} with the nearest points $x_f$ of $x'_f$ in $\hat{\mathbb{P}}$. 

\begin{rem}
Though the algorithm presented here was explained within the framework of learning optimal control with DMPs, it could be easily generalized for other frameworks that use an alternative primitive for encoding optimal solutions, for instance, Gaussian processes \citep{clever2017cocomopl}.
\end{rem}

\begin{rem}
Note that \eqref{eq:estimate_Vx}
assumes the invertibility of $g$. Nevertheless, the sampling scheme presented here can also be applied in cases where it is possible to partially recover some of the derivatives of $V$ by \eqref{eq:input_OCP_sol}, even when $g$ is not invertible. Such is the case of mechanical systems, which can be represented by \eqref{eq:affine_control_sys} and satisfy 
$$
\frac{\partial V}{\partial \dot{q}} = -2 M(q) R \tau^*,
$$
where $q$ is the vector of generalized coordinates, $M$ is the mass matrix and $\tau$ is the optimal input torque. 
\end{rem}

\renewcommand{\algorithmicrequire}{\textbf{Input:}}
\algdef{SE}[DOWHILE]{Do}{doWhile}{\algorithmicdo}[1]{\algorithmicwhile\ #1}%

\begin{algorithm*}
\caption{Sampling method for a specific direction $v$}\label{alg:sampling}
\begin{algorithmic}[1]
\Require $v$: initial direction to start sampling
\Procedure{sample}{$v$}

\State $J_\text{threshold} \gets$ threshold of allowed deviation from estimated optimal cost
\State $t_\text{samples} \gets$ maximum number of allowed samples in $\hat{\mathbb{P}}$
\State $\Delta x \gets$ size of step in direction $v$
\State $t_\text{steps} \gets$ maximum number of steps
\State $x_f \gets $ a given point in $\mathbb{P}$
\State $\hat{\mathbb{P}} \gets \emptyset$

\While{$|\hat{\mathbb{P}}| \le t_\text{samples}$}
    \State $x_f \gets x'_f$
    \State $(z^*, v^*) \gets$ solution of \eqref{eq:reverse_OCP} with $z(0) = x_f$ 
    \State $x^*(t) \gets z^*(t_f - t)$
    \State Encode optimal solution $x^*$ in $\Phi_\textrm{D}(x_f\mid x_f)$
    \State Include $x_f$ in $\hat{\mathbb{P}}$ and store weights $\omega_{i,j}$ of $\Phi_\textrm{D}(x_f\mid x_f)$ in memory
    \State $V_x \gets $ gradient of $\tilde{V}$ with respect to $x$ computed by \eqref{eq:compute_backwards_grad_V}
    \State $J_s \gets$ cost of optimal solution $v^*$
    \State $n_\text{step} \gets 1$ %
    \Do
        \State $x'_f \gets x'_f + \Delta x \cdot v$ 
        \State $\hat{V} \gets J_s + V_x \cdot x'_f$ %
        \State $J_{\textrm{DMP}} \gets $ cost of $\Phi_D(x'_f \mid x_f)$
        \State $n_\text{step} \gets n_\text{step} + 1$
    \doWhile{$J_\textrm{DMP} - \hat{V} < J_\text{threshold}$ %
            and $n_\text{step} < t_\text{steps}$ %
            and $x'_f \in \mathbb{P}$}
\EndWhile
\EndProcedure
\end{algorithmic}
\end{algorithm*}

\section{Numerical Simulation}
In this numerical simulation, we illustrate the application of the proposed method described in Algorithm~1 for the optimal control problem
\eqref{eq:OCP} with 
$ L(x(\cdot), u(\cdot)) = u^\mathsf{T} u$
and considering a non-linear system in the form \eqref{eq:affine_control_sys}, given by
\begin{equation}\label{eq:example_sys1}
    \begin{bmatrix}
        \dot{x}_1(t) \\
        \dot{x}_2(t)
    \end{bmatrix} =
    \begin{bmatrix}
         -x_1^2(t) \\
         -2x_2(t)
    \end{bmatrix}
    +
    \begin{bmatrix}
        1 &  x_1(t) \\
        0 &  1
    \end{bmatrix}
    \begin{bmatrix}
       u_1(t) \\
       u_2(t)
    \end{bmatrix}.
\end{equation}
Moreover, the boundary conditions are $x_0 = (5, 5)$ for the initial state $x(0)$, and the terminal state $x(8)$ will be equal to $x_f$, which will be a varying parameter within
$$\mathbb{P} = \{(x_1, 5) : x_1 \ \in [1, 9]\}.$$ 

\begin{rem}
We only considered variations in the $x_1$ direction since similar results would be obtained by considering variations in the $x_2$ direction. 
\end{rem}

Algorithm~1 was executed by taking $x_0$ as the first sampling point and with direction $v=(1,0)$. The threshold for the cost difference ($J_\text{threshold}$) was chosen as $10$, and the maximum number of allowed samples ($t_\text{samples}$) in this direction was chosen as $15$. The size of the step ($\Delta_x$) used was $0.2$, and the maximum number of steps ($t_\text{steps}$) allowed without taking any sample was $5$.

\begin{figure}
    \centering
    \includegraphics[scale=0.54]{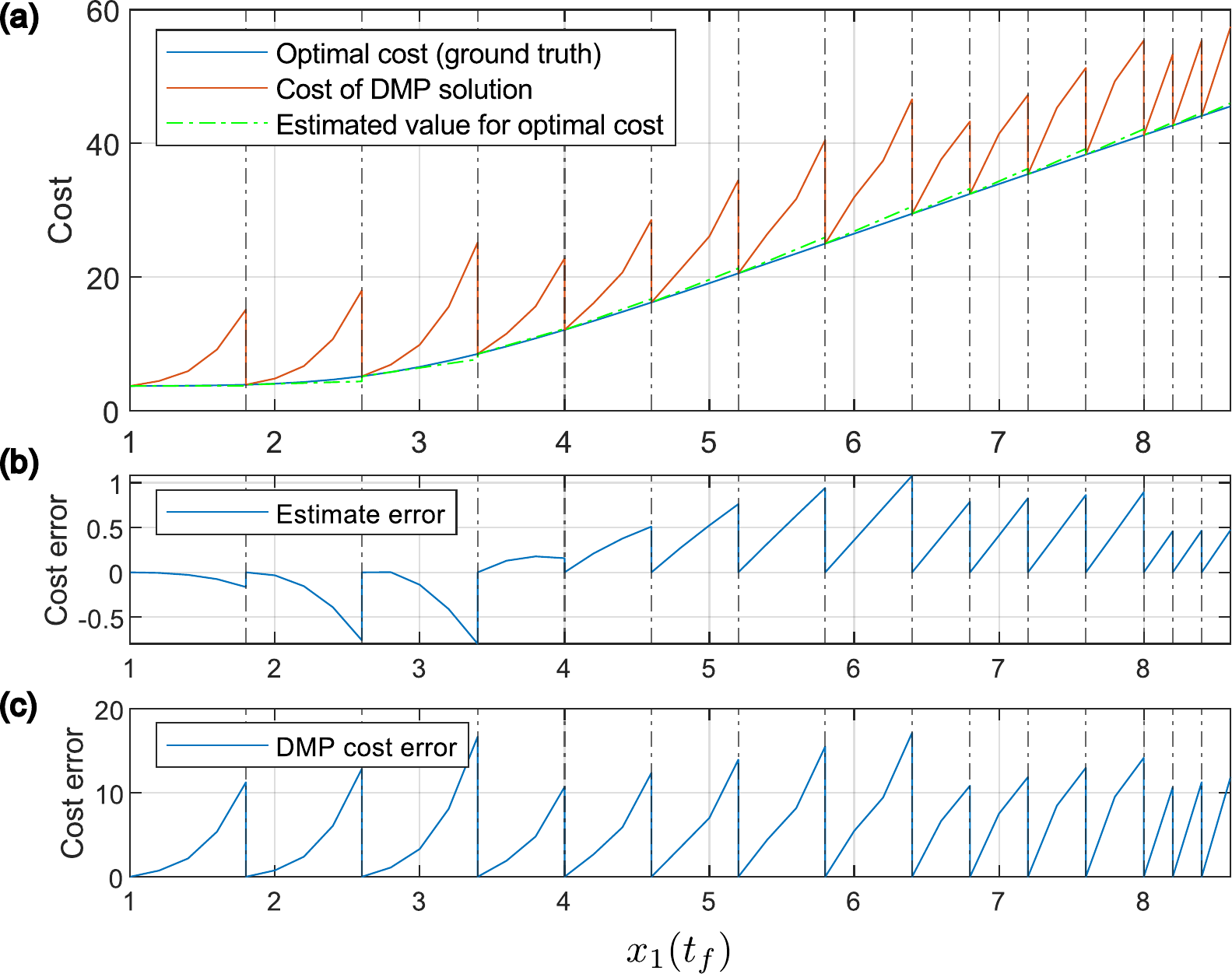}
    \caption{%
    (\textit{a}) Application of sampling algorithm to example system \eqref{eq:example_sys1}. The dashed vertical lines indicate the terminal points wherein a new DMP should be used, while the points between the lines are out-of-sample learned trajectories.  %
    (\textit{b}) Error between the estimated optimal cost and the real value of the optimal cost. %
    (\textit{c}) Difference between the cost of the DMP trajectory and the real optimal cost. %
    }\label{fig:value_estimate}
\end{figure}

Figure~\ref{fig:value_estimate} shows the efficacy of the approximation method for the optimal value function using \eqref{eq:first-order-approx}. The maximum absolute error between the optimal cost (computed using a numerical optimal control solver only for the purpose of benchmark) and the cost estimated by \eqref{eq:first-order-approx} is given by $1.07$.  

\begin{figure}
    \centering
    \includegraphics[scale=0.53]{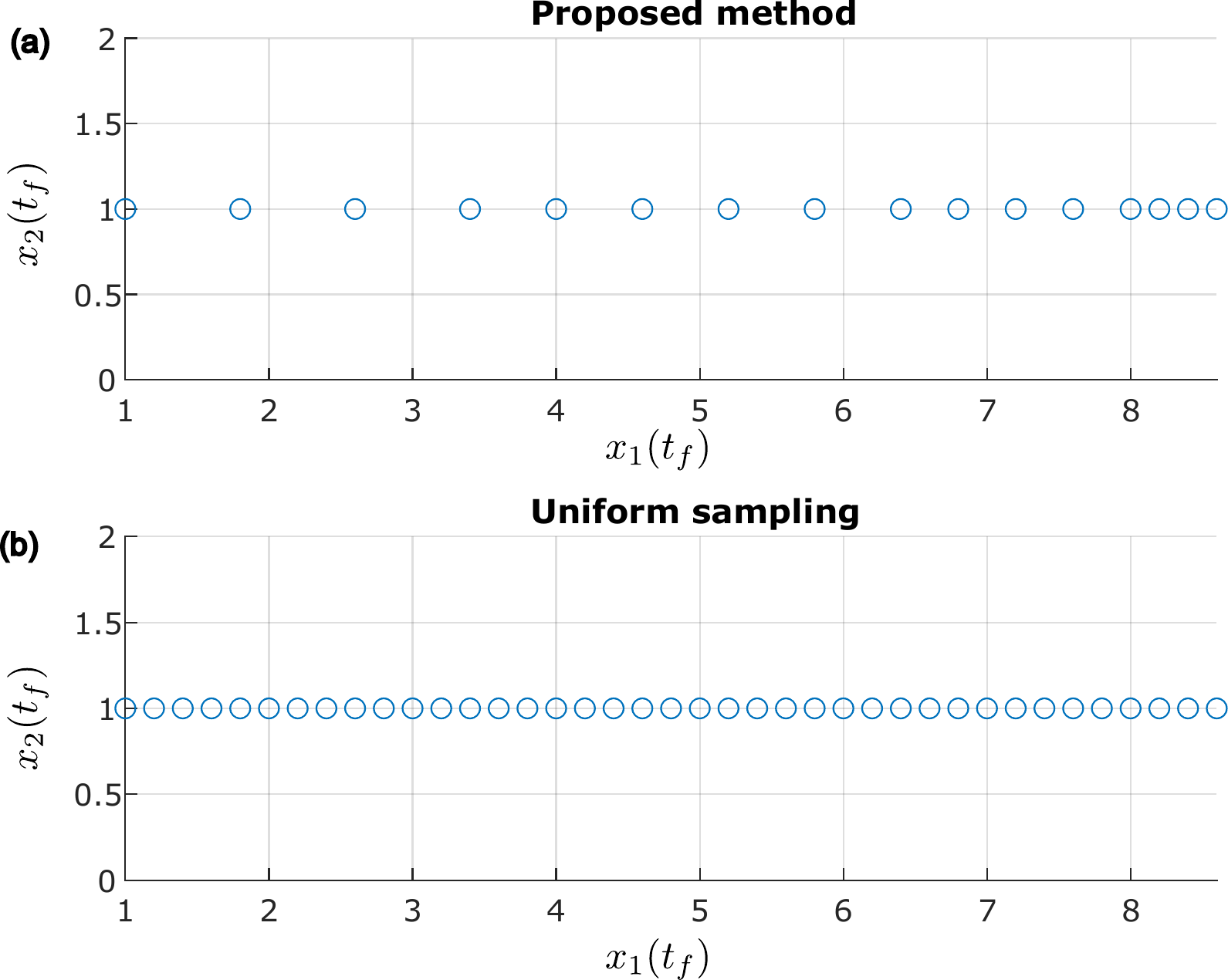}
    \caption{%
    (\textit{a}) Grid obtained by proposed Algorithm 1. It uses only $15$ samples. %
    (\textit{b}) Uniform grid using the minimal distance between sampling points of grid obtained by Algorithm 1. It uses $39$ samples.
    }\label{fig:grid}
\end{figure}

To clarify the difference between the proposed sampling strategy and the naive uniform strategy, Figure~\ref{fig:grid} shows the respective grids that are obtained by each strategy. It can be seen that the resulting grid $\hat{\mathbb{P}}$ is non-uniform, reflecting the high sensitivity of the cost when $x_f$ is within the subset $\{(x_1, 5) : x_1 \ \in [7, 8]\} \subset \mathbb{P}$. On the other hand, a uniform grid with length given by the minimal distance (of $0.2$) between the sampling points of the uniform grid would require at least $39$ samples to cover $\mathbb{\hat{P}}$, more than the double of the samples obtained using the proposed sampling strategy. 

\section{Conclusion}

In this paper, we augmented a DMP-based framework for learning optimal control with a method to estimate the optimal value near the learned optimal trajectories, and used it to derive a sampling scheme that could be used for reducing storage requirements in applications where this could be an issue, such as in embedded systems.

The results in this paper could be extended in many directions, such as how to generalize this framework for dealing with optimal control problems with other varying parameters beside the terminal state. This would be interesting, for instance, for selecting the optimal configuration of a reconfigurable robot to execute a given task or in optimal braking problems, such as the one presented in \citep{HamadEtAl2023Fastbrakingmaneuvers}.

\appendix
\section*{Appendix}

\subsection*{Proof of Proposition~\ref{thm:upper_bound_DMP_traj}}
Let $x(t) := \Phi_\textrm{D}(x_f\mid x_f)(t)$ and $x'(t) := \Phi_\textrm{D}(x_f'\mid x_f)(t)$. Then the following equations 
are valid:
\begin{align*}
-\tau^{2}\ddot{x}(t)+\kappa(x_{f}-x(t))-D\tau\dot{x}(t) & =F(s(t)), \\
-\tau^{2}\ddot{x}'(t)+\kappa(x'_{f}-x'(t))-D\tau\dot{x}'(t) & =F(s(t)).
\end{align*}
Defining $\hat{x}(t) = x(t) - x'(t)$ and subtracting the second equation from the first one yields
\begin{equation*}
-\tau^{2}\ddot{\hat{x}}(t)-\kappa \hat{x}(t)-D\tau\dot{\hat{x}}(t)=-\kappa(x_f-x_f'),
\end{equation*}
a second-order linear equation with a constant forcing term which solution (when $\kappa = D^2/4$) is given by 
$$
\hat{x}(t) = 
\left(e^{-\frac{D t}{2 \tau }} \left(-\frac{D t}{2 \tau }-1\right)+1\right) (x_f-x_f').
$$

\end{document}